\documentclass[letterpaper, 10 pt, conference]{ieeeconf}  

\IEEEoverridecommandlockouts                              

\usepackage{graphicx}
\usepackage{mathrsfs} 
\usepackage{multirow}
\usepackage{colortbl}
\usepackage{microtype}

\usepackage{comment}
\usepackage[hang,flushmargin]{footmisc}
\usepackage{booktabs}

\usepackage{amsfonts}
\usepackage{amssymb}
\usepackage{amsmath}
\usepackage{stfloats}

\definecolor{commentcolor}{RGB}{110,154,155} 

\definecolor{azure(colorwheel)}{rgb}{0.0, 0.5, 1.0}
\definecolor{nicegreen}{rgb}{0.0, 0.7, 0.1}
\definecolor{ashblue}{rgb}{0.36, 0.54, 0.66}
\definecolor{ashgrey}{rgb}{0.7, 0.75, 0.71}
\definecolor{applegreen}{rgb}{0.55, 0.71, 0.0}
\definecolor{blue}{rgb}{0.0, 0.0, 1.0}
\definecolor{postechred}{rgb}{0.784, 0.003, 0.313}
\definecolor{gu}{rgb}{0.5460, 0.1755, 0.2766}
\definecolor{jy}{rgb}{0.58, 0, 0.827}
\definecolor{ballblue}{rgb}{0.13, 0.67, 0.8}
\definecolor{cornellred}{rgb}{0.7, 0.11, 0.11}
\definecolor{darkcyan}{rgb}{0.0, 0.55, 0.55}
\definecolor{CuGray}{gray}{0.9}
\definecolor{airforceblue}{rgb}{0.36, 0.54, 0.66}
\definecolor{rev}{rgb}{0.784, 0.003, 0.313}
\definecolor{pink}{cmyk}{0, 0.7808, 0.4429, 0.1412}
\definecolor{amethyst}{rgb}{0.6, 0.4, 0.8}
\definecolor{black}{rgb}{0.0, 0.0, 0.0}
\definecolor{tb3_yellow}{rgb}{0.996, 1.0, 0.6}
\definecolor{tb3_orange}{rgb}{0.980, 0.8, 0.604}
\definecolor{tb3_red}{rgb}{0.972, 0.6, 0.6}
\definecolor{dimgray}{rgb}{0.41, 0.41, 0.41}
\definecolor{brickred}{rgb}{0.8, 0.25, 0.33}
\definecolor{bleudefrance}{rgb}{0.19, 0.55, 0.91}
\definecolor{blue(ncs)}{rgb}{0.265, 0.445, 0.765}
\definecolor{blue(ryb)}{rgb}{0.01, 0.28, 1.0}
\definecolor{orange}{rgb}{1.0, 0.49, 0.0}
\definecolor{Gray}{gray}{0.88}
\definecolor{green(ncs)}{rgb}{0.0, 0.62, 0.42}
\definecolor{brightpink}{rgb}{1.0, 0.0, 0.5}
\definecolor{midgray}{rgb}{0.55, 0.55, 0.55}

\definecolor{kellygreen}{rgb}{0.3, 0.73, 0.09}

\newcolumntype{g}{>{\columncolor{CuGray}}c}
\newcolumntype{z}{>{\columncolor{CuGray}}l}

\renewcommand{\paragraph}[1]{\noindent\textbf{#1.}\,\,}
\newcommand{\wj}[1]{\textcolor{black}{#1}}

\newcommand{\yohan}[1]{\textcolor{black}{#1}}

\usepackage{xspace}

\makeatletter
\def\@fnsymbol#1{\ensuremath{\ifcase#1\or *\or \dagger\or \ddagger\or
   \mathsection\or \mathparagraph\or \|\or **\or \dagger\dagger
   \or \ddagger\ddagger \else\@ctrerr\fi}}
\makeatother

\def\onedot{.\@\xspace}
\def\eg{\emph{e.g}\onedot} 
\def\ie{\emph{i.e}\onedot}

\newcommand{\Sref}[1]{Sec.~\ref{#1}}

\newcommand{\Fref}[1]{Fig.~\ref{#1}}

\newcommand{\be}{\begin{eqnarray}}
\newcommand{\ee}{\end{eqnarray}}
\newcommand{\bee}{\begin{eqnarray*}}
\newcommand{\eee}{\end{eqnarray*}}

\newcommand{\matrixb}{\left[ \begin{array}}
\newcommand{\matrixe}{\end{array} \right]}

\usepackage{amssymb}
\usepackage{pifont}

\usepackage{booktabs,multirow,xcolor,pifont}
\usepackage{algorithm}
\usepackage{algpseudocode}   

\usepackage{float}
\usepackage{url}

\makeatletter
\renewcommand\fs@ruled{%
  \def\@fs@cfont{\bfseries}%
  \let\@fs@capt\floatc@ruled
  \def\@fs@pre{\vspace*{6pt}\hrule height.8pt depth0pt \kern2pt}%
  \def\@fs@post{\kern2pt\hrule\relax}%
  \def\@fs@mid{\kern2pt\hrule\kern2pt}%
  \let\@fs@iftopcapt\iftrue
}
\makeatother

\usepackage{adjustbox}
\usepackage{capt-of}

\newcommand{\drop}[1]{\textsubscript{\textcolor{gray}{#1}}}

\title{\LARGE \bf
\yohan{DarkQA: Benchmarking Vision-Language Models on \\Visual-Primitive Question Answering in Low-Light Indoor Scenes}
}

\author{Yohan Park$^{1}$, Hyunwoo Ha$^{2}$, Wonjun Jo$^{2}$, and Tae-Hyun Oh$^{1}$
\thanks{Corresponding author: Tae-Hyun Oh. 
This work has been submitted to the IEEE for possible publication. Copyright may be transferred without notice, after which this version may no longer be accessible.}%
\thanks{$^{1}$Yohan Park and Tae-Hyun Oh are with Korea Advanced Institute of Science and Technology (KAIST), Daejeon 34141, South Korea
        {\tt\small (e-mail: john.a.park@kaist.ac.kr, taehyun.oh@kaist.ac.kr).}}%
\thanks{$^{2}$Hyunwoo Ha and Wonjun Jo are with Pohang University of Science and Technology (POSTECH), Pohang 37673, South Korea
        {\tt\small (e-mail: hyunwooha@postech.ac.kr, jo1jun@postech.ac.kr).}}%
}
\begin{document}



\maketitle
\thispagestyle{empty}
\pagestyle{empty}


\begin{abstract}
Vision Language Models (VLMs) are increasingly adopted as central reasoning modules for embodied agents. Existing benchmarks evaluate their capabilities under ideal, well-lit conditions, yet robust 24/7 operation demands performance under a wide range of visual degradations, including low-light conditions at night or in dark environments--a core necessity that has been largely overlooked. 
To address this underexplored challenge, we present \yohan{DarkQA}, an open-source benchmark for evaluating perceptual primitives under multi-level low-light conditions in embodied scenarios.
\yohan{DarkQA evaluates single-view egocentric observations across controlled degradation levels, isolating low-light perceptual failures before they are entangled with complex embodied tasks}.
\yohan{The benchmark contains 9.4K deterministically generated and verifiable question--image pairs spanning five visual-primitive families.} A key design feature of \yohan{DarkQA} is its physical fidelity: 
visual degradations are modeled in linear RAW space, simulating physics-based illumination drop and sensor noise followed by an ISP-inspired rendering pipeline; \yohan{we further validate the synthesis against real paired low-light camera data}. 
We evaluate representative VLMs and Low-Light Image Enhancement (LLIE) preprocessing methods. 
Results show consistent VLM degradation under low illumination and sensor noise, while LLIE provides severity-dependent but unstable recovery. 
We demonstrate the utility of \yohan{DarkQA} by evaluating a wide range of state-of-the-art VLMs and Low-Light Image Enhancement (LLIE) models, \yohan{and} systematically reveal VLMs' limitations when operating under these challenging visual conditions.
Our code and benchmark dataset will be released upon acceptance. 
Project website: \url{https://darkqa-benchmark.github.io}
\end{abstract}
\section{INTRODUCTION}


\label{sec:intro}
\yohan{Advances in vision-language models (VLMs) have significantly enhanced robotic perception and decision-making, supporting} semantic scene understanding~\cite{peng2023openscene}, spatial reasoning~\cite{yang2025thinking}, and vision-language-action (VLA) policies~\cite{zitkovich2023rt, driess2023palm}. 
However, household robots are often intended for 24/7 operation, which means they will frequently encounter low-light scenarios, such as nighttime, entering dark rooms or power blackouts.
\yohan{Such low-light conditions are especially problematic because they globally weaken the visual-primitives—the visual evidence available to VLMs to tackle embodied tasks.
In embodied pipelines, these front-end perception failures can affect downstream tasks (\eg, memory update, spatial or affordance reasoning).
}
\wj{
Representative egocentric and embodied QA benchmarks are not designed to systematically evaluate VLMs under controlled low-light visual degradation~\cite{grauman2022ego4d, datta2022episodic, jia2022egotaskqa, mangalam2023egoschema, zhou2025egotextvqa, das2018embodied, azuma2022scanqa, majumdar2024openeqa}.
}
\yohan{Illumination remains an orthogonal and under-controlled evaluation variable, even though robust perception under low illumination is not an edge case but a core necessity for real-world embodied deployment}~\cite{keller2020illumination}.
\yohan{Accordingly, benchmarks that explicitly stress-test VLMs on visual-primitives under controlled low illumination in embodied scenarios are essential to quantify visual robustness before it is confounded with navigation, memory, or action failures.}
Nevertheless, acquiring large-scale, real-world low-light images with clean, paired annotations—ideally with corresponding well-lit reference views—is challenging and costly, which has hindered the construction of such benchmarks.
As a result, existing benchmarks have largely overlooked systematic evaluation of VLM-based reasoning and perception under degraded illumination, limiting their ability to predict real-world robustness.

\begin{figure}[t]
    \centering
    \includegraphics[width=0.95\columnwidth]{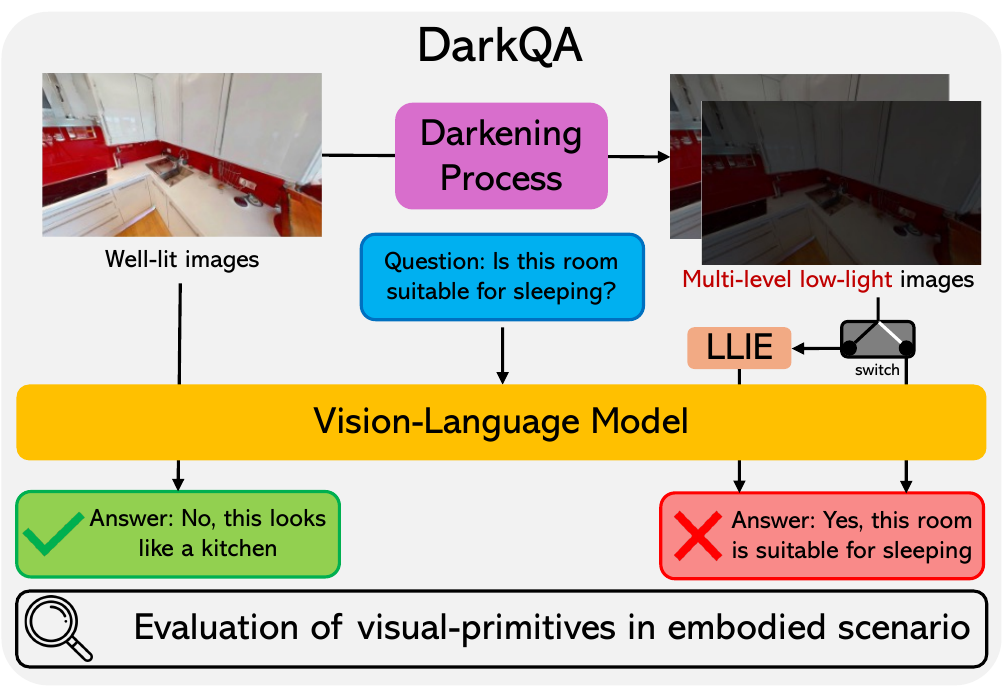}
    \vspace{-0.5em}
    \caption{\textbf{Illustration of the \yohan{DarkQA} benchmark.}
    We present \yohan{DarkQA}, a new benchmark that evaluates VLM robustness in visual-primitives under low-light conditions in embodied scenario.
    \yohan{DarkQA} assesses VLM performance under two distinct conditions: clean, well-lit inputs (L0) and a multi-level ladder of physics-based low-light images (L1-L5).
    Furthermore, the benchmark examines the effect of applying Low-Light Image Enhancement (LLIE) models as a pre-processing step.}
    \label{fig:darkqa_teaser}
    \vspace{-1.0em}
\end{figure}

To fill this evaluation void, we present \yohan{DarkQA}, an open-source benchmark \yohan{for evaluating VLM robustness in visual-primitive question answering (QA) from egocentric indoor observations across controlled low-light levels, from moderate degradation to extreme stress-test regimes}.
The design of \yohan{DarkQA} is primarily grounded in a physically based formulation, where all visual degradations are modeled at the RAW sensor data level (or in linear RGB space). This follows the physics of illumination and sensor noise to realistically simulate real-world Image Signal Processing (ISP) scenarios.
Moreover, to ensure benchmark integrity and prevent potential data contamination~\cite{shumailov2024ai}, all Question Answering (QA) pairs are deterministically generated via rule-based procedure, rather than depending on commodity VLM services.
QA generation results in a family of queries targeting perceptual primitives, including from simple object recognition (\eg, ``Is there a cushion in the image?'') to affordance reasoning (\eg, ``I want to sleep, is this room suitable for this?'').

\yohan{DarkQA} provides 9.4k question--image pairs, a standardized evaluation protocol, and a public codebase to reproduce our low-light degradation pipeline. Our \yohan{DarkQA} benchmarks a diverse set of vision--language models (VLMs), including both open- and closed-source systems~\cite{liu2024improved, li2024llava, wang2025internvl3, yang2025qwen3, hurst2024gpt}. 
We also evaluate \yohan{four low-light image enhancement (LLIE) models}~\cite{feijoo2025darkir}\yohan{~\cite{cai2023retinexformer, guo2020zero, liu2021retinex}} as preprocessing baseline\yohan{s}. 
Our evaluation yields two observations. 
First, all tested VLMs show a clear performance decline as the images degrade. 
Second, \yohan{LLIE preprocessing is method- and severity-dependent: it can improve QA accuracy at some degradation levels, but is not uniformly beneficial and does not recover well-lit performance.} 
Together, these results show that current VLMs remain brittle under low-light corruption, and that perceptual enhancement alone is insufficient as a general solution, motivating robustness-oriented evaluation and method development.

\begin{figure*}[t]
    \centering
    \adjustbox{max width=0.9\textwidth, margin=0pt 0pt 0pt 10pt}{%
        \includegraphics{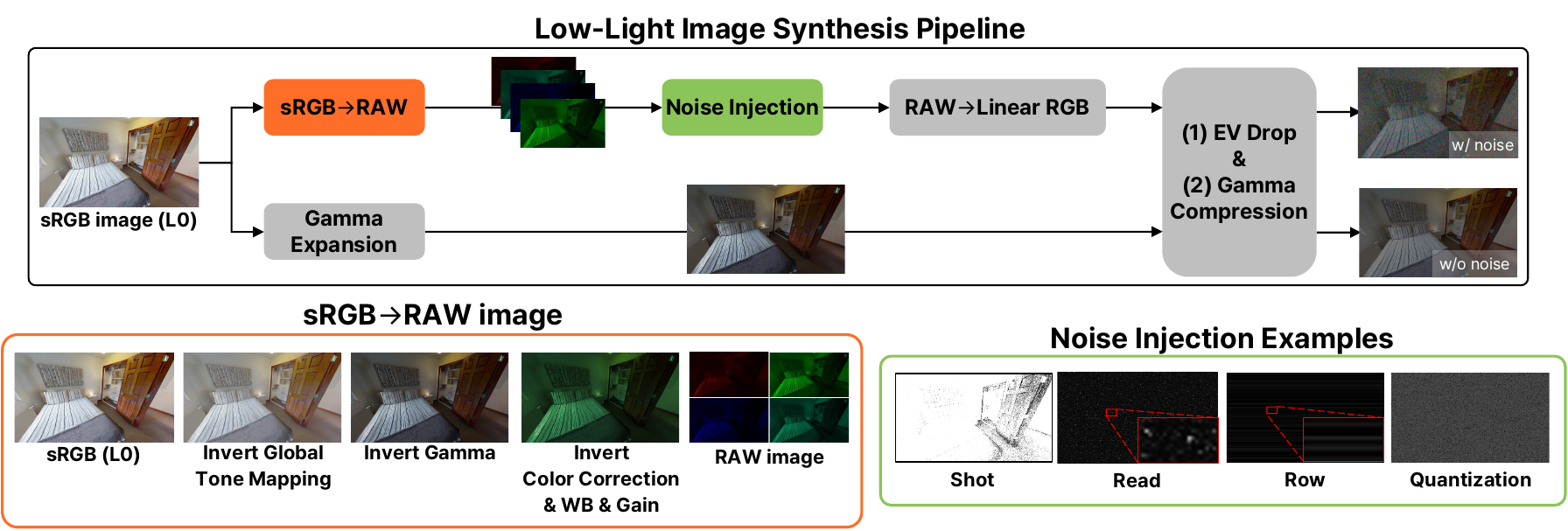}%
    }
    \caption{
    \textbf{Low-light synthesis pipeline with disentangled illumination and noise factors.}
    To generate controlled low-light inputs for our benchmark, we adopt an ISP-inspired unprocessing and noise formulation from prior work~\cite{brooks2019unprocessing,wei2021physics}. Crucially, we produce \emph{paired} variants for each original image to disentangle failure sources in VLM-based QA: (a) a physics-based branch (top) that unprocesses sRGB to Bayer RAW, injects four noise components in RAW, and then applies EV drop and gamma compression; and (b) a noise-free branch (bottom) that applies the same EV drop in linear RGB without noise injection. This paired design enables separate evaluation of performance degradation due to illumination reduction versus sensor noise. The bottom-left panel summarizes the sRGB$\rightarrow$RAW unprocessing steps, and the bottom-right panel visualizes the four noise components (shot, read, row-pattern, and quantization noise) as independent signals. The small \textcolor{red}{red boxes} in the read and row noise examples indicate zoomed-in crops for visualization.
    }
\label{fig:low_light_image_synthesis_pipeline}
\end{figure*}

\section{RELATED WORK}
\label{related_work}



\wj{
\subsection{Egocentric Question Answering Benchmarks}
Egocentric question answering evaluates visual-language reasoning from first-person observations. Prior benchmarks study large-scale egocentric video understanding~\cite{grauman2022ego4d}, episodic-memory QA~\cite{datta2022episodic}, task-level reasoning~\cite{jia2022egotaskqa}, long-form video QA~\cite{mangalam2023egoschema}, and scene-text-aware assistance~\cite{zhou2025egotextvqa}. However, none of these benchmarks evaluates VLMs under controlled dark or low-light visual degradation. Unlike prior work~\cite{zhang2025egonight}, our benchmark evaluates VLM robustness under controlled, multi-level synthesized low-light degradation.
}

\wj{
Embodied QA benchmarks such as EmbodiedQA~\cite{das2018embodied}, ScanQA~\cite{azuma2022scanqa}, and OpenEQA~\cite{majumdar2024openeqa} are closely related, as they evaluate agents that answer questions about embodied or 3D environments. Yet they focus on navigation, 3D scene QA, or environment-level memory rather than isolating low-light visual robustness. In contrast, \emph{DarkQA} evaluates egocentric indoor QA under controlled low-light levels, directly assessing VLM robustness to degraded visual-primitives.
}



\wj{
\subsection{Handling Low-Light Images}
Recent research addresses low-light visual perception in two directions.
The first improves recognition under low illumination for task-specific vision problems, such as depth estimation, object detection, or pose estimation~\cite{wang2021regularizing, lee2023human, sasagawa2020yolo}.
While effective, these studies do not evaluate VLM robustness for visual-primitive QA from egocentric indoor observations under controlled low-light degradation.
The second direction is low-light image enhancement (LLIE), which aims to improve brightness, contrast, and detail visibility for human perception or downstream models.
Representative LLIE models include DarkIR~\cite{feijoo2025darkir}, RetinexFormer~\cite{cai2023retinexformer}, ZeroDCE~\cite{guo2020zero}, and RUAS~\cite{liu2021retinex}.
Although LLIE improves visual quality, its effect on VLM-based reasoning over low-light egocentric inputs remains underexplored.
We therefore evaluate whether LLIE preprocessing helps VLMs answer visual-primitive questions in dark environments.
}
\begin{figure*}[t]
    \centering
    \adjustbox{max width=0.9\textwidth, margin=0pt 0pt 0pt 10pt}{%
    \includegraphics{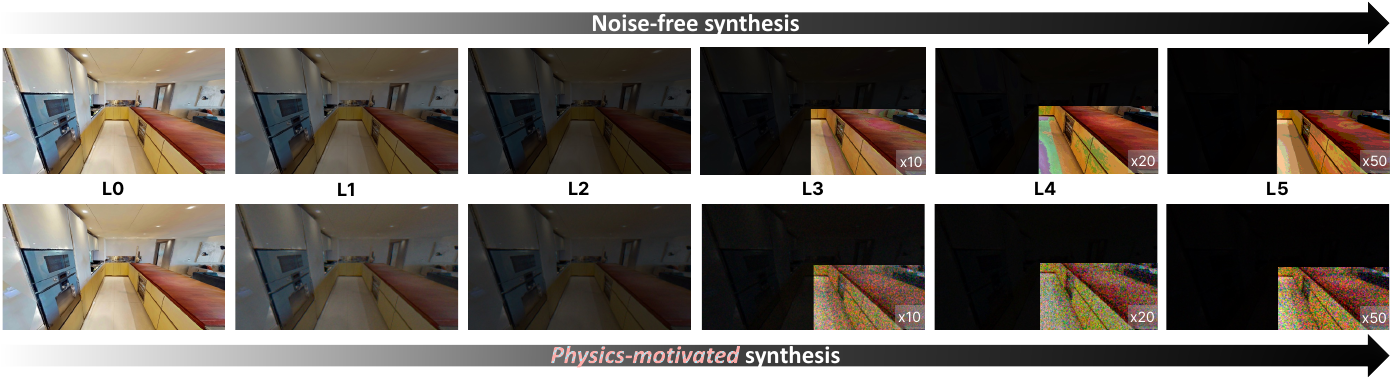}%
    }
    \caption{\textbf{Example low-light image synthesization.}
    Synthesized low-light image examples across degradation levels L0–L5. The top row shows EV drop only, while the bottom row shows EV drop combined with noise injection. The lower-right insets show $1/4$-image crops with pixel intensities amplified for visibility; the numbers (×10, ×20, ×50) indicate the amplification factor.}
    \label{fig:low_light_example}
\end{figure*}

\section{\yohan{DarkQA}: Dataset Construction \\ and QA Pair Generation}
\label{sec:method}

Our \yohan{DarkQA} is designed to evaluate VLMs’ recognition of core \yohan{visual} primitives from a single image-question pair under controlled low-light conditions. 
However, acquiring real-world low-light images with clean, paired annotations is challenging. 
To address this, we synthesize low-light images from the well-established indoor scene dataset (\ie, HM3D-Sem ~\cite{ramakrishnan2021hm3d}).
This section describes the low-light image synthesis for \yohan{VLM} input (\Sref{ssec:synthesis_low}) and the \yohan{visual-primitive QA} dataset construction process (\Sref{ssec:dataset}).
A key feature of our work is a dataset construction pipeline designed for high reproducibility and expandability. 

\subsection{Low-Light Image Synthesis for Benchmark Inputs}
\label{ssec:synthesis_low}
Low-light images suffer from two distinct physical degradations. First, the reduced photon count leads to a fundamental loss of signal, which we term illumination degradation (\ie, exposure-value (EV) drop). Second, this weakened signal yields a low Signal-to-Noise Ratio (SNR), as sensor noise (\eg, shot, read, pattern, and quantization noise) becomes dominant relative to the remaining signal~\cite{wei2021physics}. To reproduce these conditions for benchmark inputs, we design a physics-based low-light synthesis pipeline. Specifically, across multiple degradation severities (L1--L5, increasing severity), we synthesize two \emph{paired} low-light variants per original image: (i) A noise-free EV-drop variant and (ii) a physics-motivated variant with level-dependent sensor noise injection in the RAW domain, as in Fig.~\ref{fig:low_light_example}. 
This design enables disentangling the respective impacts of illumination degradation and sensor noise on perceptual peformance of VLMs. 

\subsubsection{Noise-free low-light image synthesis}
Exposure-value (EV) drop is applied at linear RGB space after decoding sRGB images as shown in the lower branch of low-light image synthesis pipeline depicted in Fig.~\ref{fig:low_light_image_synthesis_pipeline}.

\paragraph{Decoding to linear RGB}
\yohan{Since EV is physically defined with respect to scene-linear irradiance, we conduct linearization before applying exposure scaling.} 
\yohan{Hence,} we \yohan{first} approximate linearization using gamma expansion. 
Let $x_{\mathrm{sRGB}}$ represent a sRGB pixel value in an input image and $x_{\mathrm{lin}}$ its linear form. Following \cite{plotz2017benchmarking,brooks2019unprocessing}, we compute
\begin{equation}
x_{\mathrm{lin}} = \left( \max(x_{\mathrm{sRGB}}, \epsilon) \right)^{2.2},
\label{eq:linearization}
\end{equation}
where $\epsilon = 10^{-8}$ ensures numerical stability.

\paragraph{Exposure scaling} Next, let $\Delta\mathrm{EV}$ denote the absolute change in exposure value. Reducing the exposure by $\Delta\mathrm{EV}$ scales the $x_{\mathrm{lin}}$ by $2^{-\Delta\mathrm{EV}}$. The exposure-scaled pixel value is \begin{equation} x'_{\mathrm{lin}} = 2^{-\Delta\mathrm{EV}} \, x_{\mathrm{lin}}. \label{eq:ev_scaling} \end{equation}

\paragraph{Re-encoding to sRGB}
Finally, the exposure-scaled pixel value $x'_{\mathrm{lin}}$ is mapped back to sRGB via gamma encoding:
\begin{equation}
x'_{\mathrm{sRGB}} = (x'_{\mathrm{lin}})^{1/2.2}.
\label{eq:srgb_encoding}
\end{equation}
We standardize an degradation levels L1--L5 with $\Delta\mathrm{EV} \in \{2.0,4.0,6.0,7.5,9.0\}$, respectively (L0 is the original).

\subsubsection{Physics-motivated low-light image synthesis} 
\label{ssec:physics_motivated_pipeline}
We synthesize realistic low-light images using a physics-based pipeline that combines ISP inversion/forward pass~\cite{brooks2019unprocessing} and raw-domain noise modeling~\cite{wei2021physics}.
The process is shown in the upper branch of low-light image synthesis pipeline of Fig.~\ref{fig:low_light_image_synthesis_pipeline}.
\yohan{Realism of our physics-based low-light image synthesis pipeline is discussed in Sec.~\ref{ssec:realism_analysis}}.

\paragraph{Unprocessing (sRGB $\rightarrow$ RAW)} We first normalize an 8-bit sRGB image $\mathbf{I} \in \{0,\ldots,255\}^{H \times W \times 3}$, 
where $H$ and $W$ denote the image height and width, respectively, to
\[
\mathbf{I}_{\mathrm{sRGB}} = \frac{\mathbf{I}}{255} \in [0,1]^{H \times W \times 3}.
\]
To obtain a camera-linear RAW image from $\mathbf{I}_{\mathrm{sRGB}}$, we invert the ISP following~\cite{brooks2019unprocessing}. 
We denote the unprocessing operator by ${u}(\cdot)$, and express the resulting Bayer RAW mosaic as 
\begin{equation}
\mathbf{B} = {u}(\mathbf{I}_{\mathrm{sRGB}}),
\label{eq:unprocess_operator}
\end{equation}
where $\mathbf{B} \in [0,1]^{\frac{H}{2} \times \frac{W}{2} \times 4}.$
The unprocessing operator ${u}(\cdot)$ consists of five steps:
(i)~inverse tone mapping, (ii)~gamma expansion, (iii)~RGB$\rightarrow$Camera color correction with sampled matrix $\mathbf{M}_{\text{rgb}\rightarrow\text{cam}}$, (iv)~inversion of white-balance/brightness gains with highlight preservation, and (v)~mosaic extraction into RGGB Bayer representation. This restores a scene-referred signal where noise statistics are defined with respect to photon counts and sensor readout electronics, not post-ISP perceptual tone curves.

\paragraph{Noise formation in RAW}
Following the physics-based formation model of~\cite{wei2021physics}, we inject
four noise components into the camera-linear RAW signal.  
Let $\mathbf{B}$ denote the clean, mosaiced RAW image obtained from unprocessing.
After converting $\mathbf{B}$ from normalized units to the sensor’s ADU domain,
we sample a system gain $K$ log-uniformly from $[0.1, 6.0]$.
The noisy RAW image is then expressed as
\begin{equation}
\mathbf{B}_{\text{noisy}} = \mathcal{N}_4 \circ \mathcal{N}_3 \circ \mathcal{N}_2 \circ \mathcal{N}_1 (\mathbf{B}, K),
\label{eq:raw_noise_operator}
\end{equation}
where $\mathcal{N}_i$ denotes the $i$-th noise operator mapping a Bayer RAW tensor and system gain K to a Bayer RAW tensor described below.
\\
\paragraph{(1) Photon shot noise}
Photon arrival is discrete and stochastic.  
For each pixel, the number of photoelectrons $N$ follows $N \sim \mathrm{Poisson}(\lambda)$ where $\lambda$ is proportional to scene irradiance.
To simulate extreme low-light capture, we apply an ISO amplification ratio
$r \in [100, 300]$:
(i) reduce the signal by $r$ (low-light capture), 
(ii) add Poisson noise, 
(iii) amplify back by $r$ using sensor gain.
This preserves the characteristic of low-photon-count statistics while allowing
the final output brightness to be controlled independently via the EV drop.
\\
\paragraph{(2) Read noise}
Readout electronics introduce an additive noise term $N_{\text{read}}$.  
We model it using a Tukey–$\lambda$ distribution with a channel-wise DC offset
(color bias).  
The scale parameter $\sigma_{\mathrm{TL}}$ grows log-linearly with the system gain $K$:
\[
\log \sigma_{\mathrm{TL}} = a_{\mathrm{TL}} \log K + b_{\mathrm{TL}} + \epsilon,
\]
capturing the heavy-tailed distribution observed under extreme low-light~\cite{wei2021physics}.
\\
\paragraph{(3) Row noise}
Line-wise variations in the readout circuitry produce banding artifacts.
Each row $i$ receives a shared offset $n_r^{(i)} \sim \mathcal{N}(0, \sigma_r^2)$, where $\sigma_r$ also scales log-linearly with $K$.
\\
\paragraph{(4) Quantization noise}
Analog-to-digital conversion introduces rounding error $N_q$ modeled as $N_q \sim \mathcal{U}(-0.5,0.5)$, where $\mathcal{U}$ represents a uniform distribution on $[-0.5, 0.5]$, assuming a standard unit (1 ADU) quantization step.

\paragraph{Simplified ISP (RAW $\rightarrow$ sRGB)} 
Converting RAW to sRGB is an inverse operation of unprocessing:
(i) white balance with sampled gains, (ii) bilinear demosaicing from RGGB Bayer to RGB, (iii) color correction using $\mathbf{M}_{\text{cam}\rightarrow\text{rgb}}$, (iv) EV drop by $\Delta\mathrm{EV}$ in linear space (multiplying intensities by $2^{-\Delta\mathrm{EV}}$) to match the target degradation levels L1--L5, (v) gamma compression, and (vi) quantization to 8-bit sRGB.

\begin{algorithm}
\caption{Deterministic procedure for QA generation}
\label{alg:engine}
\begin{algorithmic}[1]
\Require Scene set $\mathcal{S}$; frames $\mathcal{F}_s$ for each $s \in \mathcal{S}$
\Ensure QA pairs $\mathcal{Q}$ with ground-truth answers
\State \textbf{Definitions:}
\State $\Omega_f = \{1,\dots,W\}\times\{1,\dots,H\}$: Pixel grid of frame $f$
\State $M_i^f \in \{0,1\}^{H\times W}$: Mask for segment $i$ in frame $f$
\State $\mathcal{A}_i \in \mathbb{R}^d$: Attribute vector for segment $i$ (semantic class, color, depth, area, bbox)
\State $\Phi_f = \{\mathcal{A}_i\}_{i=1}^{N_f}$: Frame statistics (all segment attributes)
\State $r_f$: Room type label for frame $f$
\State $\mathcal{C}_f \subseteq \{1,2,3,4,5\}$: Viable question families of frame $f$
\State
\State \textbf{Generate QA from Frames}
\State $\mathcal{Q} \gets \emptyset$
\For{$f \in \bigcup_{s \in \mathcal{S}} \mathcal{F}_s$} \Comment{Process each frame exactly once}
\State \textbf{--- 1. Extract Statistics}
\State Load $I_{\text{RGB}}^f, I_{\text{depth}}^f, I_{\text{sem}}^f, I_{\text{over}}^f$
\For{$i \in \text{Segments}(I_{\text{over}}^f)$}
    \State $M_i^f(x,y) \gets \mathbf{1}[(x,y)\in\Omega_f \wedge I_{\text{over}}^f(x,y)=i]$
    \State $\mathcal{A}_i \gets \text{ComputeStats}(M_i^f, I_{\text{RGB}}^f, I_{\text{depth}}^f, I_{\text{sem}}^f)$
\EndFor
\State $\Phi_f \gets \{\mathcal{A}_i : \forall i\}$ \Comment{Collect stats for frame $f$}
\State \textbf{--- 2. Generate QA}
\State $r_f \gets \text{ClassifyRoom}(\Phi_f)$
\State $\mathcal{C}_f \gets \text{Survey}(\Phi_f, r_f)$ \Comment{Find viable question types}
\For{$k \in \mathcal{C}_f$} \Comment{Generate all viable questions}
    \State \yohan{$($}q\yohan{$,a)$} $\gets \text{Rule}_k(\Phi_f, r_f)$
    \State $\mathcal{Q} \gets \mathcal{Q}\; \cup$ \yohan{$($}q\yohan{$,a)$}
\EndFor
\EndFor
\State \Return $\mathcal{Q}$
\end{algorithmic}
\end{algorithm}

\begin{figure}[t]
    \centering
    \adjustbox{max width=\linewidth, margin=0pt 0pt 0pt 8pt}{%
        \includegraphics{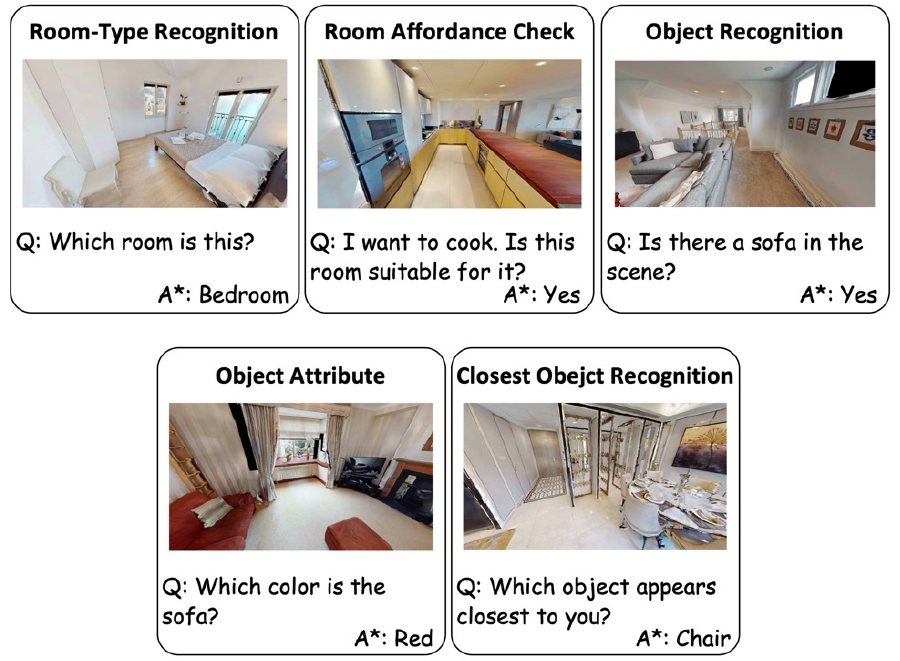}%
    }
    \caption{\textbf{Question family of our \yohan{DarkQA} benchmark.}
    Five \yohan{DarkQA} question categories with examples. \yohan{DarkQA} encompasses questions asking room-type recognition, room affordance check, object recognition, object attribute.}
    \label{fig:question_family}
\end{figure}

\subsection{Dataset Construction}
\label{ssec:dataset}
We build the dataset for evaluation upon a representative subset of 52 scenes from HM3D-Sem~\cite{ramakrishnan2021hm3d}, selected for diversity and semantic richness. 
For each scene, we record a human-demonstrated navigation trajectory that systematically explores the environment to maximize spatial coverage. 
To generate the ground-truth QA pairs, we uniformly subsample the trajectory and select keyframes at a fixed time interval (\eg, one frame every 2,s), rendering their geometric and semantic modalities (\eg, RGB, depth, segmentation). 
We then use Algorithm~\ref{alg:engine} as deterministic procedure to automatically generate QA pairs from the \yohan{common interface $(I_{\mathrm{RGB}}^f \in \mathbb{R}^{H \times W \times 3}, I_{\mathrm{depth}}^f \in \mathbb{R}^{H \times W}, I_{\mathrm{sem}}^f \in \mathbb{R}^{H \times W}, I_{\mathrm{over}}^f \in \mathbb{R}^{H \times W})$ (RGB image, depth map, semantic label map, and over-segmentation map, respectively.), which can be applied to any dataset that can be mapped to the interface for each frame.}
This approach ensures each question has a single verifiable answer by filtering ambiguities (\eg, tiny objects), requires no manual \yohan{QA} annotation \yohan{beyond the source annotations}, and avoids potential data contamination by not using commodity VLM services. 
This entire process is fully reproducible.

Algorithm~\ref{alg:engine} operates in two stages: frame-statistics extraction (Stage~1) and QA generation (Stage~2). 
In Stage~1, we cache the frame statistics $\Phi_f$ required for Stage~2. 
Each frame $f$ is represented as a quadruple 
$f = (I_{\mathrm{RGB}}^f, I_{\mathrm{depth}}^f, I_{\mathrm{sem}}^f, I_{\mathrm{over}}^f)$. 
\yohan{For each segment mask $M_i^f$ obtained from $I_{\mathrm{over}}^f$, 
$\mathrm{ComputeStats}(M_i^f, I_{\mathrm{RGB}}^f, I_{\mathrm{depth}}^f, I_{\mathrm{sem}}^f)$ 
extracts an attribute vector $\mathcal{A}_i$ summarizing the segment's information.
The collection of all segment attributes forms the frame statistics 
$\Phi_f=\{\mathcal{A}_i\}_{i=1}^{N_f}$.}

\yohan{In Stage~2, $\mathrm{ClassifyRoom}(\Phi_f)$ assigns a room label $r_f$ using deterministic object-signature rules, such as beds for bedrooms, toilets or showers for bathrooms, etc.
Given $\Phi_f$ and $r_f$, $\mathrm{Survey}(\Phi_f,r_f)$ selects the viable question families $\mathcal{C}_f$, and each $\mathrm{Rule}_k:(\Phi_f,r_f)\mapsto(q,a)$ instantiates the corresponding question template and ground-truth answer, where $k$ indexes the five families in Fig.~\ref{fig:question_family}.
A rule is applied only when its required evidence is available and unambiguous in $\Phi_f$.
}
For example, consider the ``Closest Object Recognition'' question in Fig.~\ref{fig:question_family}. Object-level statistics are first extracted. The QA generation pipeline validates two conditions: (i) at least two non-structural, non-quasi-2D object instances with valid depth measurements exist, and (ii) the depth gap between top-two closest objects exceeds a minimum threshold to ensure perceptual validity. If satisfied, the closest object is determined as the ground-truth answer. In this example, ``chair'' is identified as the closest object.

This pipeline generates five question families targeting visual-primitives for embodied operation: \emph{Room-Type Recognition}, \emph{Room Affordance Check}, \emph{Object Recognition}, \emph{Object Attribute}, and \emph{Closest Object Recognition}. The examples for each family are provided in Fig.~\ref{fig:question_family}. \yohan{These categories are chosen to cover atomic visual evidence commonly required for embodied operation. Evaluating these primitives separately isolates low-light perceptual failures before they are entangled with complex downstream embodied tasks.}

\subsection{Dataset Statistics}
\label{ssec:dataset_statistics}
\begin{figure}[t]
    \centering
    \adjustbox{max width=\linewidth, margin=0pt 0pt 0pt 8pt}{%
        \includegraphics{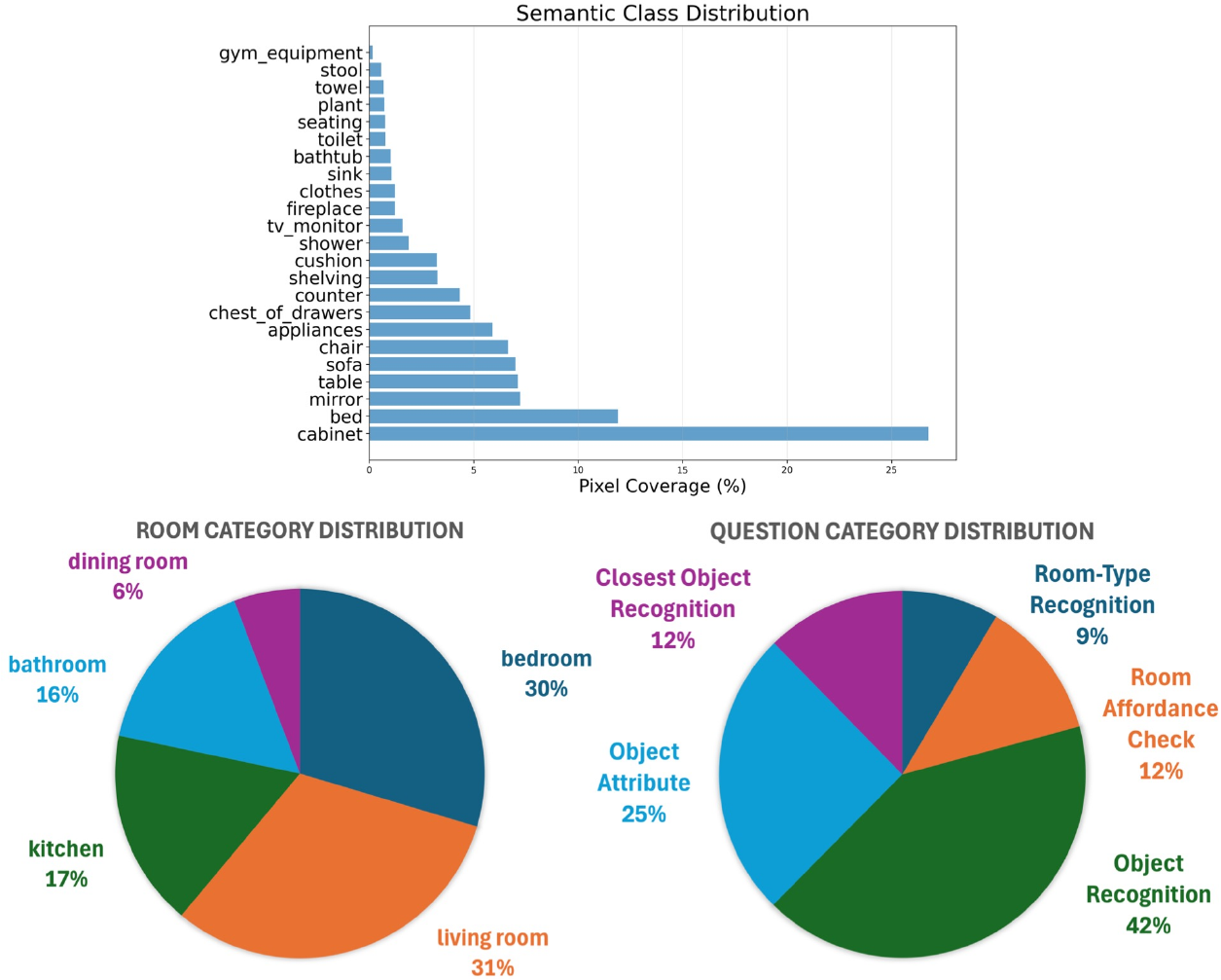}%
    }
    \caption{\textbf{Statistics of our \yohan{DarkQA} benchmark.}
    Dataset statistics, including semantic-class coverage, room-category distribution, and question-category distribution.
    }
    \label{fig:bench_stat}
\end{figure}

Our \yohan{DarkQA} comprises 52 scenes selected from HM3D-Sem, yielding 3,911 frames at $1440 \times 2560$ resolution with $\sim$9.4K QA pairs. Fig.~\ref{fig:bench_stat} shows that the dataset exhibits semantic class and room category distributions that are representative of typical residential environments. The semantic annotation covers 23 non-structural object classes, with the most prevalent being cabinet, bed, mirror, and table taking up about 53\%. Room category distribution reflects the natural spatial composition of household scenes. The question distribution across the five question families shows moderate imbalance, with frequencies determined by the geometric and semantic constraints of our rule-based QA generation pipeline and subsequent validation through human sanity checks to ensure answer correctness.

\section{EXPERIMENTS}

\label{sec:experiments}
In this section, we describe our experimental settings and provide quantitative evaluation results of various VLMs on \yohan{DarkQA, along with VLMs performance on DarkQA, validation of physics-motivated low-light synthesis pipeline, and open-ended question anwering evaluation.}

\subsection{Experimental Setup}
We evaluate \yohan{DarkQA} on both VLMs and text-only LLMs (blind LLMs). For each keyframe and degradation condition, we present a single question together with a fixed, small set of candidate answers (room-type labels, object classes, color names, or a candidate list for closest objects). VLMs receive the image and the question–choice template, whereas blind LLMs see only the textual question and choices.
Each question is thus cast as a multiple-choice problem, and models are instructed to output exactly one answer from the choices. This constrains the response space, avoids ambiguities in free-form generation, and enables exact-match scoring.

\subsection{Baseline Models}
\paragraph{Blind LLMs} We set the scenario of blind agents that produces an answer based on the question that requires visual information to answer~\cite{majumdar2024openeqa}. Even though our \yohan{DarkQA} focuses on the VLM's behavior according to illumination change and noise injection, we use the result of blind LLMs to catch the possible bias of our dataset while also testing how well the questions may be answered with an assumption of indoor environments. For the LLM choice, we report the results of GPT-4~\cite{achiam2023gpt} and LLaMA-3.1-8B~\cite{dubey2024llama}.

\paragraph{VLMs}
We evaluate a range of VLMs across different parameter scales. For 7–8B models, we report results for LLaVA-1.6-7B~\cite{liu2024improved}, LLaVA-OneVision-8B~\cite{li2024llava}, InternVL3.5-8B~\cite{wang2025internvl3}, and Qwen3-VL-8B~\cite{yang2025qwen3}. For larger-scale models (\ensuremath{\geq} 30B), we additionally evaluate InternVL3.5-30B\cite{wang2025internvl3} and Qwen3-VL-32B~\cite{yang2025qwen3} using the same respective series. 
Finally, we include GPT-4o~\cite{hurst2024gpt} as \yohan{an} upper bound.


\wj{\paragraph{LLIE model}
For VLM evaluation on LLIE-enhanced low-light images, we use DarkIR~\cite{feijoo2025darkir}, RetinexFormer~\cite{cai2023retinexformer}, ZeroDCE~\cite{guo2020zero}, and RUAS~\cite{liu2021retinex}.
}

\begin{figure*}[!t]
    \centering
    \adjustbox{max width=0.95\textwidth, margin=0pt 0pt 0pt 10pt}{%
    \includegraphics{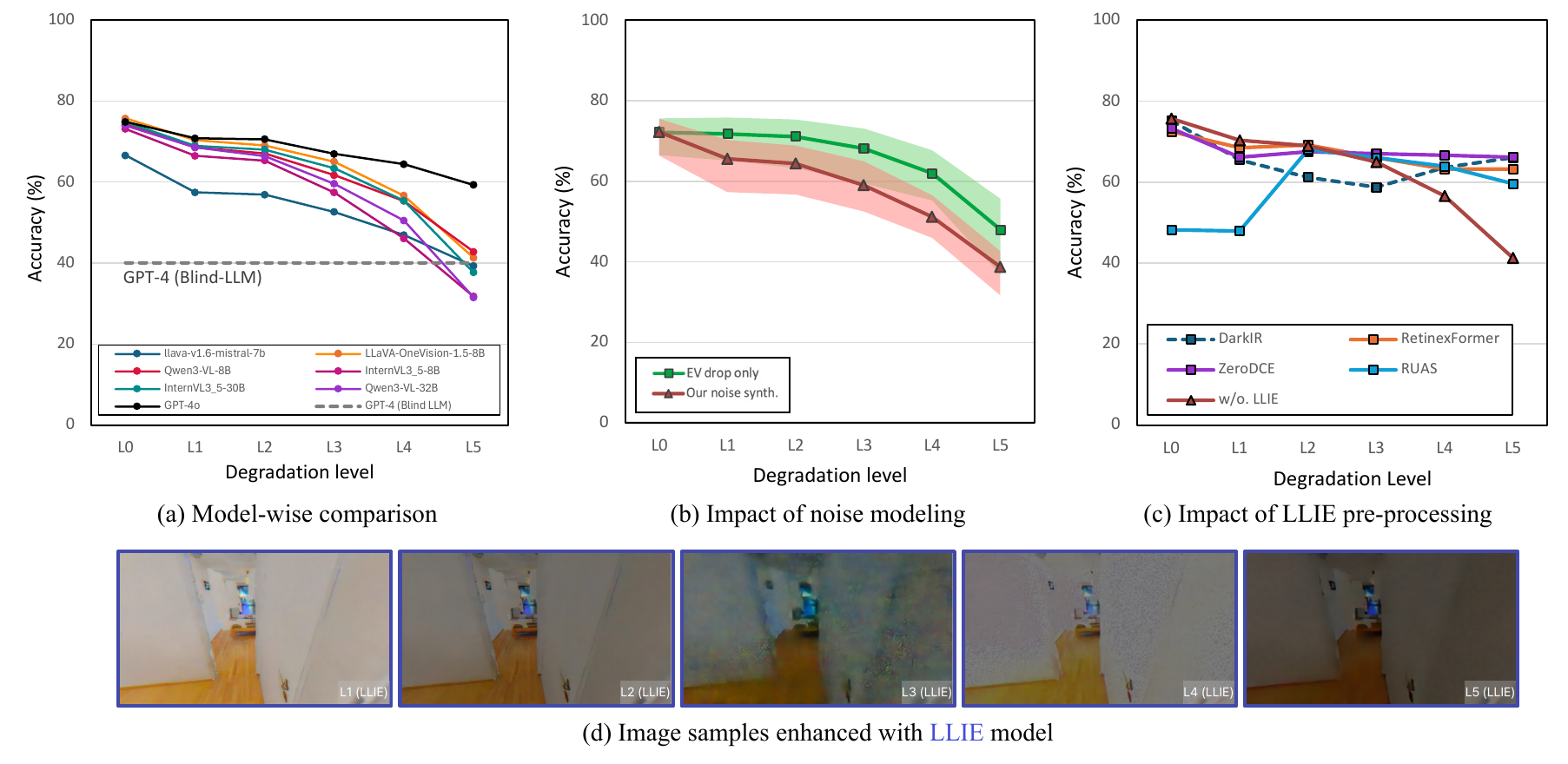}%
    }
    \caption{\textbf{Summary of the evaluation results on our \yohan{DarkQA}.}
    \emph{Degradation level} indicates the severity of low-light corruption: $L0$ corresponds to the original (well-lit) input, and higher levels ($L1 \rightarrow L5$) denote progressively darker (lower-illumination) inputs.
    We evaluate a range of open-source VLMs (LLaVA~\cite{liu2024improved,li2024llava}, InternVL~\cite{wang2025internvl3}, and Qwen-VL~\cite{yang2025qwen3} series, 7B--32B). The shaded regions in (b) denote the minimum–maximum accuracy across models at each degradation level.
    \yohan{\textbf{(a) Model-wise comparison. }} 
    \yohan{\textbf{(b) Impact of noise modeling. }}
    \yohan{\textbf{(c) Impact of LLIE pre-processing. }}
    \wj{\textbf{(d) Image samples enhanced by the DarkIR~\cite{feijoo2025darkir} model.}}
    We include GPT-4 as a Blind-LLM baseline (evaluated without vision; \textcolor{gray}{gray dashed line}) and GPT-4o~\cite{hurst2024gpt} as an upper-bound reference (black line).}
\label{fig:results_analysis}
\end{figure*}

\begin{figure}[!t]
    \centering
    \includegraphics[width=0.9\columnwidth]{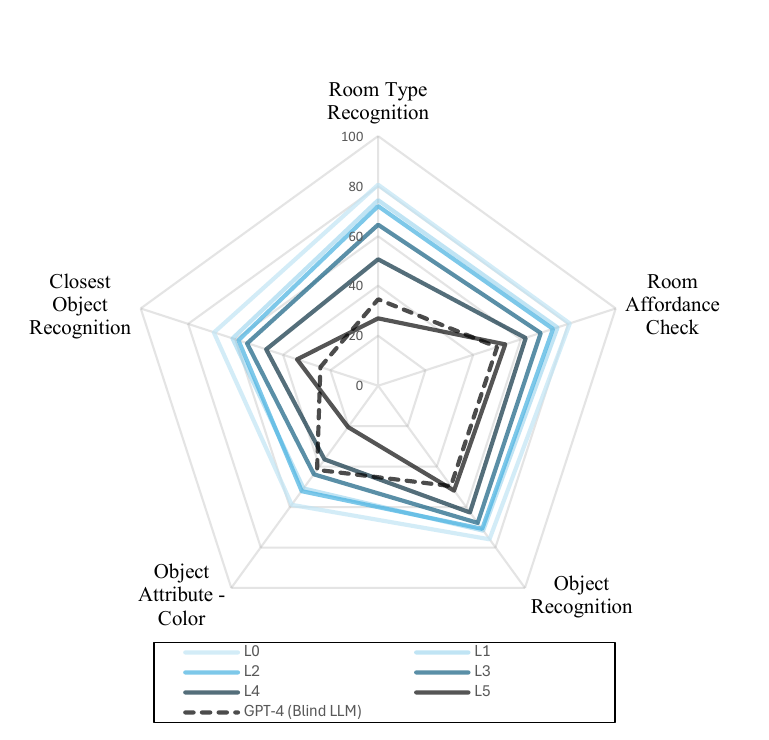}
    \caption{\textbf{Question-wise accuracy.} We plot VLM accuracy across different question types under increasing low-light degradation, where darker lines indicate more severe degradation and the \textcolor{gray}{gray dashed line} denotes the GPT-4 Blind-LLM baseline.
We observe significant drops in ``Room Type Recognition'' and ``Object Attribute – Color,'' where VLM performance falls below the GPT-4 Blind-LLM baseline.}
    \label{fig:question_wise}
\end{figure}

\subsection{VLMs performance on DarkQA}

\paragraph{Model-specific accuracy}
\label{ssec:model_specific_analysis}
Fig.~\ref{fig:results_analysis}-\yohan{(a)} provides a detailed comparison of the performance trends across individual VLMs under noisy inputs without LLIE preprocessing. While the specific degradation curves vary slightly across each models, the overall trend is a largely similar decline in accuracy as low-light conditions intensify. Although the commodity service GPT-4o consistently demonstrates the highest performance, it also shows performance degradation under low-light conditions. Furthermore, we observe an interesting point: at the most severe low-light level (L5), some VLMs achieve accuracy lower than that of GPT-4 (Blind-LLM baseline), which operates solely on textual input without any visual information. This indicates that for images under extreme degradation, the models are unable to effectively utilize these visual information, leading to a poorer understanding of semantic information compared to relying purely on language priors.

\yohan{To contextualize the severe-degradation results, we additionally compute a random-choice baseline on the full DarkQA evaluation set. The overall chance accuracy is 33.35\%, due to the mixed candidate-set sizes across question families. Thus, model accuracies around 30--35\% at L4/L5 indicate near chance-level performance, and should not be interpreted as evidence of successful visual understanding.}

\paragraph{Impact of illumination drop and sensor noise}
\label{ssec:degradation_analysis}
To understand the robustness of VLMs against visual illumination degradation, we first observe their performance under two types of low-light simulation: (1) pure EV drop and \yohan{(2) physics-motivated noise modeling}.
As shown in \Fref{fig:results_analysis}-\yohan{(b)}, both degradations consistently lead to a significant decrease in VLM accuracy. Notably, the introduction of sensor noise compounds this decline, resulting in a more pronounced performance drop compared to pure EV reduction. This confirms that VLMs are indeed highly sensitive to such visual degradation, with noise being a critical factor.

\paragraph{Effectiveness of low-light image enhancement (LLIE) pre-processing}
\label{ssec:llie_effectiveness}
Given the observed performance degradation, we investigate whether pre-processing low-light images with a state-of-the-art Low-Light Image Enhancement (LLIE) model~\cite{feijoo2025darkir} can mitigate these issues. We apply LLIE models to the noise-added low-light images before feeding them into the VLMs. As illustrated in Fig.~\ref{fig:results_analysis}-\yohan{(c)}, this approach yields mixed results. While we observe a significant accuracy improvement at more severe low-light levels (L4 and L5), performance decreases at moderate levels (L1--L3). This unstable behavior highlights the challenge of reliably enhancing low-light images across different levels of degradation. While current LLIE models enhance perceptual quality, the results suggest that current LLIE models may be biased to certain degradation levels as in Fig.~\ref{fig:results_analysis}-(d).

\paragraph{Question-wise analysis}
\label{ssec:question_wise_analysis}
To gain a more granular understanding of the performance decline, we further analyze the accuracy degradation across different question types, as shown
in Fig.~\ref{fig:question_wise}. While most categories exhibit a steady decline, we observe a critical phenomenon in two specific types: ``Room Type Recognition'' and ``Object Attribute - Color''. For these categories, the VLM accuracy drops below that of the GPT-4 (Blind-LLM) baseline at severe degradation levels (L5 for the former, and L4 and L5 for the latter). The fact that this effect is particularly pronounced for the ``Color'' category strongly suggests that VLMs struggle to extract or preserve essential visual semantic information, such as color, when processing heavily dark images.
Interestingly, this observation is analogous to the behavior of the human vision in dark scenes, where the visual primarily relies on rod cells that are sensitive to luminance because color-sensitive cone cells function much less effectively.

\yohan{\paragraph{VLMs' Hallucination and Bias}
We find interesting results on Object Recognition errors under severe low-light degradation. Interestingly, L5 does not increase object-presence false positives; instead, the model becomes more conservative, with a false-positive rate of only 4.96\% and a false-negative rate of 89.36\%. In other words, the dominant failure is missing present objects, rather than hallucinating absent ones. 
}

\yohan{For non-binary questions, severe darkness also induces answer collapse: at raw L5, 67.0\% of Room Type predictions are ``living room'' although it accounts for only 32.4\% of ground-truth room labels, and 87.7\% of Color predictions become ``black'' although black accounts for only 15.0\% of ground-truth colors. 
These analyses are based on LLaVA-OneVision-1.5-8B performance on DarkQA.
}

\yohan{\subsection{Realism of physics-motivated low-light synthesis pipeline}
\label{ssec:realism_analysis}
To validate whether our phyiscs-motivated synthetic degradations reflect real low-light camera observations, we conduct a quantitative paired-image validation using all paired normal-/low-light images in the validation split of LLRGBD-real~\cite{zhang2022lisu}, which contains real indoor images captured by an Intel RealSense D435i RGB-D camera. For each normal-light image, we synthesize its low-light counterpart using either a na\"ive EV-drop-only baseline or our physics-motivated pipeline with RAW-domain sensor-noise injection, while keeping the EV reduction identical, and compare both against the corresponding real low-light image. As shown in Table~\ref{tab:realism_lowlight_synthesis_comparison}, our pipeline improves image similarity from 25.39/0.757 to 26.06/0.785 in PSNR/SSIM and, more importantly, reduces the Kullback--Leibler divergence between real and synthesized noise-residual distributions from 14.8 to 3.3, following the residual-distribution validation protocol used in real-noise synthesis~\cite{fu2023srgb}. These results indicate that the proposed RAW-domain noise model better matches real low-light camera statistics than simple brightness reduction, thereby mitigating the real-to-sim concern while preserving the controllability required for systematic benchmark construction.}

\begin{table}[t]
\centering
\caption{\yohan{
Quantitative realism validation of \\low-light synthesis variants on paired \\real normal-/low-light images from LLRGBD-real~\cite{zhang2022lisu}. }}
\label{tab:realism_lowlight_synthesis_comparison}
\begin{tabular}{lccc}
\toprule
Method & PSNR $\uparrow$ & SSIM $\uparrow$ & $D_{\mathrm{KL}} \downarrow$ \\
\midrule
Na\"ive EV-only baseline & 25.39 & 0.757 & 14.8 \\
Ours                     & 26.06 & 0.785 & \textbf{3.3} \\
\bottomrule
\end{tabular}
\end{table}

\wj{
\begin{table}[t]
\centering
\caption{
Open-ended OpenEQA EM-EQA evaluation under our low-light synthesis pipeline, with a human visibility study on degraded observations from 32 HM3D scenes. Scores are LLM-Match (\%).
}
\label{tab:openeqa_lowlight_stress_human_compact}
\small
\begin{tabular}{lccc}
\toprule
\textbf{Method} & \textbf{L0} & \textbf{L2} & \textbf{L4} \\
\midrule
GPT-4o 
& 75.0
& 72.4 {\drop{-2.6}}
& 66.8 {\drop{-8.3}} \\
Human
& 85.1
& 74.2 {\drop{-10.9}}
& 50.7 {\drop{-34.4}} \\
\bottomrule
\end{tabular}
\end{table}
}

\vspace{-4mm}
\wj{
\subsection{Open-Ended Embodied Question Answering Evaluation}
\label{sec:openeqa_lowlight}
We further validate whether the observed low-light vulnerability extends beyond our visual-primitive QA setting at embodied question answering using the HM3D subset of OpenEQA EM-EQA~\cite{majumdar2024openeqa}.
We keep the original scenes, trajectories, viewpoints, questions, and answers fixed, apply our physics-based low-light synthesis only to the RGB observations in episodic memory, and evaluate GPT-4o with the original LLM-Match protocol.
}

\wj{
This experiment is intended as a complementary validation. DarkQA is designed as a controlled visual-primitive QA benchmark, but OpenEQA allows us to check whether the same low-light degradation trend also appears in an open-ended episodic-memory QA format without predefined answer choices. Since all non-illumination factors are kept fixed, the consistent drop in LLM-Match suggests that low-light degradation affects not only our controlled visual-primitive QA setting, but also broader embodied QA settings built on episodic visual observations.
}

\wj{
Table~\ref{tab:openeqa_lowlight_stress_human_compact} shows that GPT-4o drops from 75.0\% at L0 to 66.8\% at L4, while human performance from 32 HM3D scenes drops from 85.1\% to 50.7\%.
This result supports our main finding: low-light degradation harms not only visual-primitive QA, but also open-ended embodied QA.
}

\section{CONCLUSION}
\label{sec:conclusion}
We introduce \yohan{DarkQA, a new benchmark designed to address an overlooked and critical regime in VLM evaluation: the lack of systematic analysis for embodied scenario in low-light conditions. Using a physically-grounded low-light image synthesis pipeline, we create a reproducible benchmark to measure VLM robustness against realistic visual degradations. }
Our findings reveal that current VLMs are brittle in the dark, and that seemingly straightforward solutions like LLIE pre-processing can yield unstable results.
\yohan{
While our benchmark reveal the vulnerabilities of VLMs to low-light conditions, a detailed failure analysis remains a valuable direction.}










\bibliographystyle{ieeetr}
\bibliography{IEEEabrv,root}

\end{document}